\documentclass[10pt,twocolumn,letterpaper]{article}

\usepackage{titlesec}
\titlespacing{\paragraph}{0pt}{*1}{*1}
\usepackage{cvpr}
\usepackage{times}
\usepackage{epsfig}
\usepackage{graphicx}
\usepackage{amsmath}
\usepackage{amssymb}

\usepackage{enumitem}
\usepackage{booktabs}

\usepackage[colorlinks,citecolor=blue,breaklinks=true,bookmarks=false]{hyperref}

\cvprfinalcopy 

\def\cvprPaperID{3837} 

\usepackage{ifthen}
\newcommand{\showComments}{no}
\newcommand{\note}[2]{
  \ifthenelse{\equal{\showComments}{yes}}{\textcolor{#1}{\sf\footnotesize{}#2}}{}
}

\begin{document}

\title{EDF:  Ensemble, Distill, and Fuse for Easy Video Labeling}

\author{Giulio Zhou\\
Carnegie Mellon University\\
{\tt\small giuliozhou@cmu.edu}
\and
Subramanya R. Dulloor\\
ThoughtSpot, inc.\\
{\tt\small dulloor@thoughtspot.com}
\and
David G. Andersen\\
Carnegie Mellon University\\
{\tt\small dga@cs.cmu.edu}
\and
Michael Kaminsky\\
Intel Labs\\
{\tt\small michael.e.kaminsky@intel.com}
}

\maketitle

\begin{abstract}
    We present a way to rapidly bootstrap object detection on unseen videos
    using minimal human annotations. We accomplish this by combining two
    complementary sources of knowledge (one generic and the other specific)
    using bounding box merging and model distillation.
    The first (generic) knowledge source is obtained from ensembling pre-trained
    object detectors using a novel bounding box merging and confidence
    reweighting scheme.
    We make the observation that model distillation with data augmentation can
    train a specialized detector that outperforms the noisy labels it was
    trained on, and train a \emph{Student Network} on the ensemble detections
    that obtains higher mAP than the ensemble itself.
    The second (specialized) knowledge source comes from training a
    detector (which we call the \emph{Supervised Labeler}) on a labeled
    subset of the video to generate detections on the unlabeled portion.
    We demonstrate on two popular vehicular datasets that these techniques work to emit bounding boxes for all vehicles in the frame with higher mean average precision (mAP)
    than any of the reference networks used, and that the combination of ensembled and human-labeled data produces object detections
    that outperform either alone.
\end{abstract}

\section{Introduction}

Advancements in machine learning, and specifically deep learning, have
brought image and video analysis nearly within the grasp of non-experts.
Unfortunately, while applying off-the-shelf models produces impressive
results, there remains a gap between the performance of a generic model
and one that has been specialized for the specific domain of interest
using fine tuning or transfer learning~\cite{transferlearning2010}.
Re-training a pre-trained model into a domain, of course, requires
a sufficient quantity of labeled data~\cite{transferlearning2014}, and producing accurate labels
is costly and time-consuming, particularly when there are privacy or
expertise constraints on the data.

In this paper, we present an approach for offline object detection
in video that (1) obtains better results
using pre-trained object detectors through a combination of model
ensembling and distillation~\cite{hinton2015distilling}; and (2) can incorporate small quantities
of provided labels to increase accuracy.  Notably, our approach does
\emph{not} attempt to create a better, general object detector;  instead,
it capitalizes on the redundancy and specificity of individual video
streams to create a specialized, trained detector that empirically performs
better \emph{on that specific video}.

To do so, we present a new approach to ensembling object detections.
Unlike classification, multiple object detectors may output non- or
partially-overlapping bounding boxes, potentially different object categories,
and different confidence values.  Our merging strategy
(Section~\ref{sec:design})
combines these factors into a single output bounding box whose behavior
matches the intuitive expectations for an ensemble:  When more models agree,
the output has higher confidence;  a single model disagreeing does not cause
the location of the bounding box to be considered erroneous.

We then use the ensembled results (i.e., predicted labels on the video we are analyzing)
to train a Student Network. Particularly when used with data augmentation,
this Student Network is able to emit object detections (again on that video)
that outperform the ensemble itself.
We evaluate our technique on both a stationary video dataset (the DETRAC
dataset, surveillance videos across multiple cameras in Beijing and Tianjin)
and a first-person moving automotive dataset (KITTI, driving videos taken from
the same car within the city of Karlsruhe).  Our
approach produces a mean average precision (mAP) that is 1.25 and 1.21x higher
than the best performing member of the ensemble.

\section{Background and Related Work}
\paragraph{Object Detection:}
The object detection literature is extensive.
Modern CNN object detection architectures
integrate feature generation, proposal, and classification into a single pipeline,
and generally fall into one of two categories:
(a) single-stage architectures (such as SSD~\cite{liu2016ssd} and YOLO~\cite{redmon2017yolo9000})
which directly predict the coordinates and class of bounding boxes, and
(b) two-stage detectors (such as Faster R-CNN~\cite{ren2015faster})
that refine proposals produced by a region proposal network.
Single-stage detectors are less expensive, but also less accurate
than their two-stage counterparts.
Other work examines how to build more effective multi-scale convolutional
feature hierarchies, such as Feature Pyramid Networks~\cite{lin2017feature} (FPN).
Our work uses the outputs of an ensemble of CNN-based object detectors
trained on Microsoft COCO~\cite{lin2014microsoft} to generate labels with
which we fine-tune a compact single-stage object detector (MobileNet-SSD FPN).

Object detection in the video setting can leverage frame-to-frame flow
estimates~\cite{zhu2017flow}
or correlation and regression-based tracking~\cite{feichtenhofer2017detect} to
detect objects more consistently across frames.
For simplicity, we run and train only per-frame object detectors but could
easily train student detectors that incorporate convolutional or flow features
across frames.

\paragraph{Model Distillation} trains a (typically smaller) student model
to match the outputs of a teacher model~\cite{hinton2015distilling}.
Distillation for classification typically trains on a weighted average of
hard labels and the teacher's softmax
outputs, often softened using a temperature $T$ to $\frac{exp(y_i / T)}{\sum_j exp(y_j) / T}$.
Here, we train a specialized detector using model distillation, which aims to
match the teacher's bounding box confidences and coordinates on the dataset of
interest.
Our process for training a custom detector bears many similarities to
techniques commonly used for model distillation, such as confidence-based
loss reweighting~\cite{furlanello2018born} and treating the teacher's outputs
as soft targets~\cite{chen2017learning}.
The way that our work differs is in how we obtain labels:
rather than simply using the detections of a single teacher detector, we use
various forms of detector ensembling to generate better labels with more
reliable confidence estimates.

\paragraph{Learning from Noisy Labels:}
One way to reduce the amount of human annotation
needed to train accurate models do so is to build models that can
learn from noisy labels, which can be generated more quickly and automatically
than clean annotations.
Approaches to learning from noisy labels generally fall into two
categories~\cite{veit2017learning}.
The first aims to learn directly on the noisy labels, focusing on noise-robust
algorithms and label cleaning procedures. The second approach is based
on semi-supervised learning techniques that propagate labels from the clean part of a
dataset onto the noisy or unlabeled portion.

Similar to our work, Tripathi et al.~\cite{tripathi2016context} train what
they call a ``pseudo-labeler'' on a labeled subset of a video to generate
bounding box labels on the unlabeled portion; this is an identical setup to our
Supervised Labelers, only their labeled subset is an inherent part of their
dataset. They use an RNN to smooth the generated bounding boxes
using only high-level category labels and a smoothness constraint. They observe
that the pseudo-labeler can sometimes produce erroneous labels, which causes
their method to fail. Our Supervised Labelers suffer from a similar problem, which
we overcome by counterbalancing their detections with ensemble detections
that are more ``generically'' robust.

Misra et al.~\cite{misra2015watch} iteratively train SVM-based detectors
starting with a small collection bounding box labels. But because they did not
have the benefit of using pre-trained detectors, they achieve a much lower
mAP score.
Snorkel~\cite{ratner2017snorkel} addresses the related problem of automating
label generation for weakly supervised learning by combining many weak
user-defined heuristics using a probabilistic model.
Our work instead uses pre-trained object detectors to generate the initial
set of noisy labels, and achieves high accuracy with minimal additional work.
Because we start with pre-trained detectors as a starting point, we have
a higher baseline for accuracy than much existing work.

\paragraph{Object Detector Ensembling:}
In single object detection, ensembling can be as simple as naively
averaging the predictions from multiple detectors. OverFeat~\cite{sermanet2013overfeat}
uses a custom bounding box merging procedure to produce localization maps for a
single object. In general, however, the single object assumption is not realistic
for most real-world video.
The authors of ResNet~\cite{he2016deep} provide a method for ensembling multiple object detectors
(with the same Faster R-CNN architecture) that processes the union over all
region proposals using an ensemble of per-region classifiers. This
approach, however, requires that all the detectors have a region proposal
mechanism and that they share the same architecture. YOLO~\cite{redmon2016you}
proposes a simple scheme whereby it screens the outputs of Fast-RCNN for
false positives.

\paragraph{Minimizing the number of labels needed to analyze new video:}
Few-shot detection aims to adapt a detector to a target domain using few target
domain examples.
A recent approach~\cite{chen2018lstd} fine-tunes a detector using a small labeled subset
of the data while using regularization to ensure that fine-tuned detector
does not diverge too much from its starting checkpoint. Although this few-shot
approach is
appropriate for detection in images, it is limited by the small number of
starting annotations. In contrast, the distillation approach used in this
paper leverages the redundancy in video to, in effect, automatically 
augment the results of the initial detectors using additional unlabeled
video.

Active learning is another technique to reduce human annotation costs by exploiting
the inherent redundancy of video~\cite{vondrick2013efficiently}.
It attempts to sample a sparse set of
keyframes for the user to label (based on some proxy for the complexity of
scene and object appearance changes) and then perform interpolation on the
annotated bounding boxes.
The end goal, however, is to perfectly annotate all frames of a video, so
label sampling occurs over fairly short time horizons, and thus requires a fairly
substantial number of labels. However, active learning techniques for choosing which
frames to label can complement our work.

\begin{figure*}[ht!]
    \centering
    \begin{minipage}{.75\textwidth}
    \includegraphics[width=0.95\textwidth]{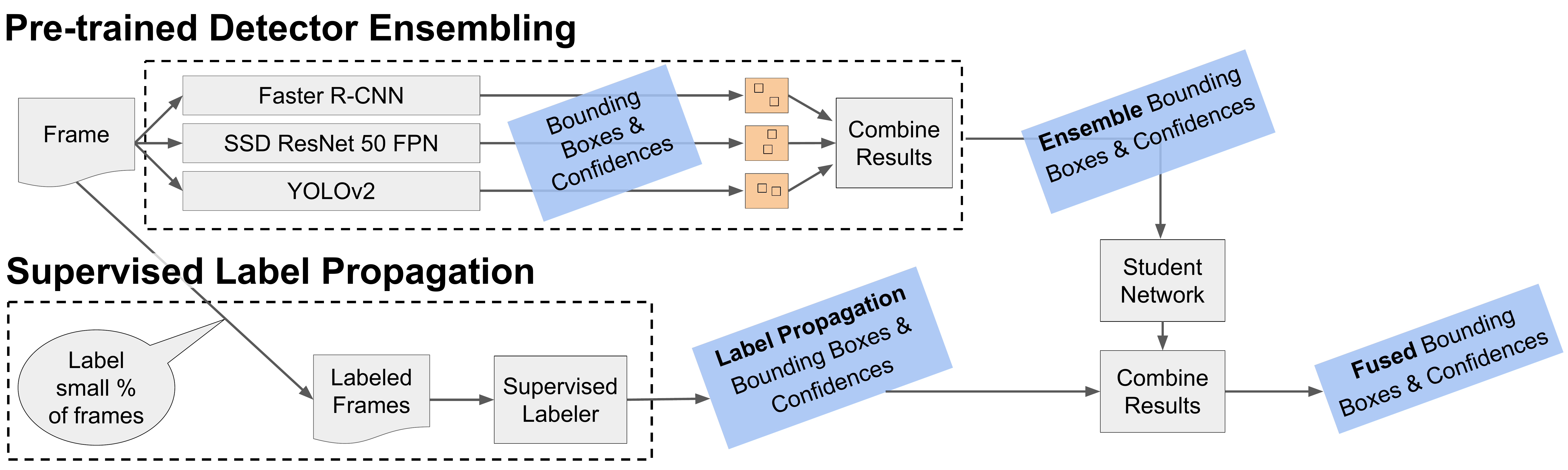}
    \caption{The training flowchart.}
    \label{fig:training_loop}
    \end{minipage}%
    \begin{minipage}{.25\textwidth}
    \centering
    \includegraphics[width=0.95\columnwidth]{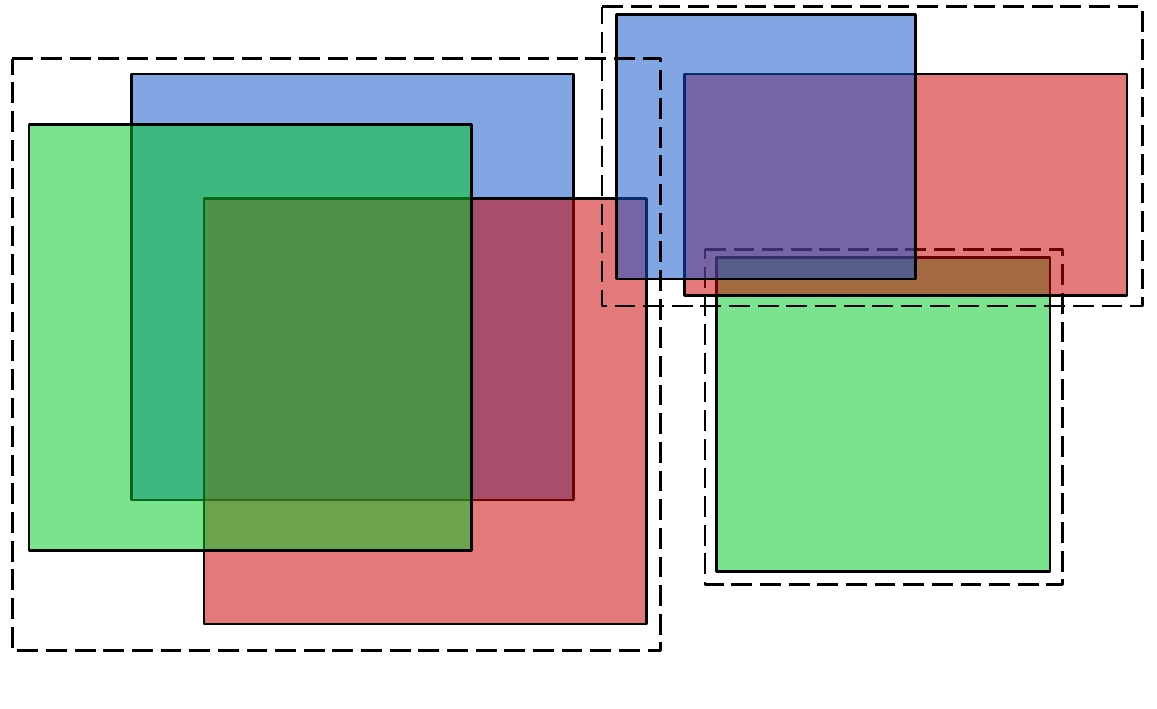}
    \caption{The ensemble bounding box merging procedure. Different colors
             indicate that the boxes were sourced from different detectors.
             On the left, the green bounding box acts as the anchor between
             the red and blue ones. On the right, the green bounding box
             is not included due to insufficient IOU with the other boxes.}
    \label{fig:merging_strategy}
    \end{minipage}
\end{figure*}

\paragraph{Video stream specialization:}
Existing work~\cite{kang2017noscope,hsieh2018focus,lu2018accelerating}
seeks to reduce the time to run expensive CNNs on video by
learning cheap video-specific and camera view specific models.
One shortcoming is that these systems are concerned only with matching the predictions of
a reference CNN rather than improving ground truth accuracy.
In addition, they often
fail to show significant improvements for moving cameras.
We believe that this area
of research is complementary to our work and could reduce the overall run time of
per-frame detectors where applicable (e.g. static video surveillance scenes).

\section{Design}
\label{sec:design}

Figure~\ref{fig:training_loop} shows an overview of our approach.  The top half
of the figure shows the basic design, where we take a set of pre-trained object
detectors and combine them into an ensemble.  We use the ensemble to produce
a new set of predictions on the target video, and use those predictions to
train a Student Network. This part of the system assumes no
access to ground truth labels.  The bottom half of the figure shows how our
system can incorporate a small number of human-provided ground truth labels to
further improve the object detection results.  We describe each part of the
design below.

\subsection{An ensemble of pre-trained object detectors}
\label{ensembling}

Ensembling is a well-known technique to improve predictions, but most of
the existing work focuses on image classification.  Applying ensembling to
object detection, however, introduces additional questions not answered
by previous work, such as how to combine spatially-scoped bounding boxes,
classifications, and confidences.  We show that successfully combining the
results from individual detectors can help overcome their individual biases
and spurious failures; this sections presents our design; the results are in
Section~\ref{results}.

To ensemble the object detectors, one must merge
bounding boxes that may only partially overlap (or may not overlap at all)
and produce a single ensemble output confidence, based on both the confidence
of the individual ensemble members as well as the overlap between the predicted
boxes.

\paragraph{Merging bounding boxes:}
Our ensemble consists of three pre-trained object detectors: Faster R-CNN, SSD
FPN Resnet-50, and YOLOv2.  We run each detector on every frame of the video.
We then apply a greedy IOU-based merging strategy (illustrated in
Figure~\ref{fig:merging_strategy}) to merge the ensemble outputs
into a discrete set of bounding boxes. This procedure is as follows:

\begin{enumerate}
    \item \emph{Match bounding boxes to their neighbor with highest IOU:}
        First, we compute the IOU between each detector's bounding box and the boxes
        generated by the other detectors. Then, we remove boxes with IOU less than
        $\mathit{IOU}_\mathit{thresh}$ from consideration.  Our evaluation,
        described below, uses a threshold of 0.7.  For a given bounding box,
        its nearest neighbors are the overlapping bounding boxes for each of
        the other detectors with the highest ($\mathit{IOU}_\mathit{thresh}$)\@.
        In our system, a bounding box could have between 0 and 2 nearest neighbors.
    \item \emph{Find ``mutually nearest tuples'':} For each bounding box,
        create a tuple consisting of itself (acting as the ``anchor'') and all of
        its nearest neighbors where the nearest neighbor relationship is
        mutual.
    \item \emph{Output merged bounding boxes:} Iterate through the mutually
        nearest tuples in descending order of cardinality (i.e., start with bounding
        boxes that have two mutually nearest neighbors). If none of its elements
        have been consumed, output merged bounding box and add elements to
        consumed set.  Merge the boxes by averaging their coordinates and
        confidences.
    \item \emph{Remove overlapping boxes using non-maximal suppression (NMS):}
        NMS (using the same $\mathit{IOU}_\mathit{thresh}$ as
        Step 1) helps reduce unmerged false positives.
\end{enumerate}

Because the pre-trained detectors we use already apply NMS with thresholds
between 0.5 and 0.6, we set $\mathit{IOU}_\mathit{thresh}$ to a somewhat
higher value (0.7) so that detections must have above-average overlap in order to be matched.

\paragraph{Combining confidences:}
The next step is to modify the bounding box confidences using detector
consensus. Without further processing, simply averaging the detection
confidences undervalues detections that have high agreement.
Intuitively, having multiple detectors agree on a detection should make
it more likely to be correct.
To reflect this increased confidence, we modify a bounding box with average
confidence $\hat{y}$ from $n$ detectors using the equation
$$\hat{y}^* = \hat{y}^\frac{1}{\beta^{(n-2)}}$$
where $\beta >= 1$. We are interested in how $n$ deviates from the ``mean
number of detections''; here, we do not modify the confidence of merged
detections from two sources.
Setting $\beta = 2$ squares the confidence for 1 detector
(decreasing its confidence) and takes the square root for 3 detectors
(increasing its confidence), while setting $\beta = 1$ ignores
the ensemble agreement altogether.
Increasing $\beta$ pushes detection confidences asymptotically closer to 0 and 1.
In our experiments, we use $\beta = 3$, and we observe little change in
confidences from setting $\beta > 3$.
Table~\ref{table:ensemble_conf_reweight} shows how increasing $\beta$
increases mAP\@.

\begin{table}
\begin{small}
    \centering
    \begin{tabular}{lcc}
        \boldmath$\beta$ & \textbf{KITTI} & \textbf{DETRAC} \\
        \midrule
        1	& 0.401 & 0.419 \\
        1.5	& 0.437 & 0.481 \\
        2	& 0.447 & 0.504 \\
        3	& 0.451 & 0.518 \\
        ~\\
    \end{tabular}
    \caption{Increasing $\beta$ upweights triplet detections and downweights
             single detections, improving mAP.}
    \label{table:ensemble_conf_reweight}
\end{small}
\end{table}

\begin{table}
\begin{small}
    \centering
    \begin{tabular}{lcc}
        \textbf{Settings} & \textbf{KITTI} & \textbf{DETRAC} \\
        \midrule
        Ensemble (2-of-3)   & 0.436 & 0.471 \\
        Ensemble (1-of-3)   & 0.401 & 0.419 \\
        Reweighted Ensemble & 0.451 & 0.518 \\
        ~\\
    \end{tabular}
    \caption{Ensemble (2-of-3) simply drops all single unmatched detections.
             Ensemble (1-of-3) keeps all detections. Reweighting uses the
             detections from Ensemble (1-of-3).}
    \label{table:all_conf_reweight}
\end{small}
\end{table}

Adjusting the confidences not only improves mAP but also improves the quality of
the labels used in training (see below).
Simply dropping all singleton detections (we term this ``Ensemble (2-of-3)'')
improves the base accuracy (Table~\ref{table:all_conf_reweight}),
but it is over-aggressive at discarding detections that may have value for training.
Reweighting instead retains low confidence detections that would otherwise be dropped
altogether:  they remain as weak positives instead of being turned into
difficult (and possibly false) negatives.

\paragraph{Distilling to a Student Network:}

To further improve detection accuracy, we use the ensemble outputs as
detections to train a Student Network (Figure~\ref{fig:training_loop}).
We show in Section~\ref{results} that this Student Network outperforms the
ensemble itself.
The Student Network is initialized using a MobileNet-SSD object detector with a
Feature Pyramid Network (FPN) base, pretrained on MS-COCO\@.
This model is relatively lightweight compared to the ensemble detectors, which
reduces its capacity to overfit.
As shown in Figure~\ref{fig:ablation_studies}, training a Student Network
improves mAP (compared to the ensemble output) from 0.451 to 0.454 on KITTI
and from 0.518 to 0.530 on DETRAC.
Adding data augmentation (specifically, random cropping) further improves the
mAP to 0.504 on KITTI and 0.554 on DETRAC\@.

The Student Network is trained on all of the video frames using the ensemble detections
as its labels.
We initially trained the Student Network using the noisy ensemble outputs as ground
truth, but found that this resulted in poor accuracy.
This approach failed because it treated high-confidence true positive
detections as equal in importance to low-confidence false positives.
We overcame this problem by adopting a technique from model distillation:
treat these ensemble detections as ``soft targets'' and directly
regress to their confidences using cross entropy loss.
This target encourages the Student Network to match the ensemble confidences directly and
leads to better generalization.

\subsection{Incorporating human labels}
The second part of our design allows the system to
accommodate a small amount of labeled data.  We use these ground truth labels
to train a Supervised Labeler using transfer learning.  Here, we
define a single label as all of the bounding boxes in a given image.  In our
experiments, we use a variable, small fraction of the dataset’s ground-truth
labels to train a Supervised Labeler, and in Section~\ref{subsec:distill}
evaluate the performance of the Student Network and the Supervised Labeler alone,
and then the combination of the two detectors.

In contrast to the Student Network, the Supervised Labeler is trained on
the small subset of frames that have clean labels.
We then use the Supervised Labeler to generate labels for the remaining
frames.

Together, the Student Network and Supervised Labeler produce two complementary sets
of detections
that can be combined to obtain better accuracy than either detector alone.
Empirically, we find that the Student Network, trained using labels from an
ensemble of
generic object detectors, is more consistent from frame-to-frame but does a poor
job of detecting objects in context (e.g., an occluded parked car). On the other
hand, the Supervised Labeler is able to account for dataset or human labeling
biases but may produce bizarre and spurious detections on unseen frames.

Finally, we combine detections from both sources to obtain improved accuracy
beyond that of either source. We do so by applying the same ensemble merging
procedure described in Section~\ref{ensembling} on the outputs of both
detectors, combining detection pairs that are mutually nearest neighbors.
We operate on the principle that the detections from the Student Network and
Supervised Labeler are
likely to disagree in terms of localization, and so use a much lower
$\mathit{IOU}_\mathit{thresh}$ of 0.3.
After merging detection coordinates and confidences, we downweight the
confidence of unmatched bounding boxes by a factor of $0.5$.

\section{Evaluation}
\label{results}

\paragraph{Datasets:}

We evaluate our system on two datasets: KITTI~\cite{geiger2013vision}
and DETRAC~\cite{wen2015ua}.
Because we focus on offline video analysis, we are not concerned with
generalizing beyond the given video collection. We, therefore,
conduct all of our experiments on the training set alone, and ignore the
provided ground truth labels except in those experiments where we train
a Supervised Labeler\@.
The end goal is to produce detections
on the given video collection that are close to the provided ground truth on the
same video collection.

The KITTI training set consists of 8,008 frames of 10 fps video divided among
21 videos from
a car driving through Karlsruhe, Germany. There are 31,790 bounding boxes among
the vehicle classes (car, truck, van). This dataset is especially challenging
because one must detect objects under heavy occlusion and in changing lighting
conditions while the object is moving (which precludes the use of techniques specific to
stationary cameras). To simplify evaluation, we remove video 0017, a short
145 frame video and the only one that does not contain any vehicles.

The DETRAC training set contains 83,791 frames of 25 fps video across 60 videos
from a
collection of surveillance cameras in Beijing and Tianjin. The four vehicle
classes (car, truck, bus, van) have 594,555 bounding boxes in total.
Performing well on this dataset requires detecting objects at multiple scales
under occlusion as well as generalizing across multiple camera views under
different weather conditions (i.e., sunny, cloudy, night, rainy).

\paragraph{Experiment Setup:}
We use the Tensorflow Object Detection API~\cite{huang2017speed} as (a) the source of two pre-trained
object detectors (Faster R-CNN and SSD FPN ResNet-50) and (b) a training
platform for fine-tuning a MobileNet-SSD detector. We use a Keras implementation
of YOLOv2~\cite{www-yolo-keras-port}, which is based on the original DarkNet
model.

We use mAP@0.5:0.95 (hereafter mAP) from COCO\@ as our accuracy metric. This metric
takes the average precision at all IOU thresholds between 0.5 and 0.95 spaced
at increments of 0.05 and averages them. Throughout the paper, we define
\emph{accuracy} as the mAP
achieved when evaluating detections obtained by processing a
given dataset against the ground truth on that same dataset.

As described previously, we merge all vehicle categories (e.g. car, truck, bus,
van) into a single vehicle superclass.  We do so because the categories defined
in KITTI and DETRAC do not overlap exactly with COCO vehicle classes---and
even for categories with the same label name (e.g., car, truck), the labeling
decisions in COCO are not necessarily consistent with the target datasets.
For example, pickup trucks and similar car-truck hybrids may not be identically
placed into the respective car or truck categories.
Class confusion is a large source of error in
object detection that we do not address in this paper. Rather, we focus on
accurately detecting all vehicles so that users can partition classes as they
choose after the fact.

We run the pre-trained detectors at the resolution that performs best for
Faster-RCNN\@. For DETRAC, this is the original 960x540 resolution.
For KITTI, this is 1920x600 (upscaled from the original resolution of 1242x375).
We train Student Networks on both datasets at the original resolutions.

We trained our student detectors on Google Cloud Platform using Cloud
TPUs. All training is performed at full (32 bit floating point)  precision.
Because TPUs do not support Non-Maximal Suppression, the only change from
standard MobileNet-SSD training is that we do not use Online Hard Example
Mining~\cite{shrivastava2016training}, which we found has little impact on
results.

\subsection{Baselines and Pre-trained Detector Accuracy}

\begin{table}
\begin{small}
    \centering
    \begin{tabular}{lcc}
        \textbf{Detector} & \textbf{KITTI} & \textbf{DETRAC} \\
        \midrule
        Faster R-CNN     & 0.416 & 0.433 \\
        SSD ResNet50 FPN & 0.422 & 0.458 \\
        YOLOv2           & 0.367 & 0.406 \\
        Ensemble         & 0.451 & 0.518 \\
        ~\\
    \end{tabular}
    \caption{Pre-trained detector mAP scores.}
    \label{table:off_the_shelf_baseline}
\end{small}
\end{table}

\begin{table}
\begin{small}
    \centering
    \begin{tabular}{lcc}
        \textbf{Settings} & \textbf{KITTI} & \textbf{DETRAC} \\
        \midrule
        Ensemble & 0.451 & 0.518 \\
        Mask R-CNN & 0.440 & 0.517 \\
        Supervised Labeler & 0.625 & 0.693 \\
        ~\\
    \end{tabular}
    \caption{The Supervised Labeler uses 3\% of labels from KITTI and
             0.5\% of labels from DETRAC.}
    \label{table:mask_rcnn}
\end{small}
\end{table}

\begin{figure}
    \centering
    \includegraphics[width=0.95\columnwidth]{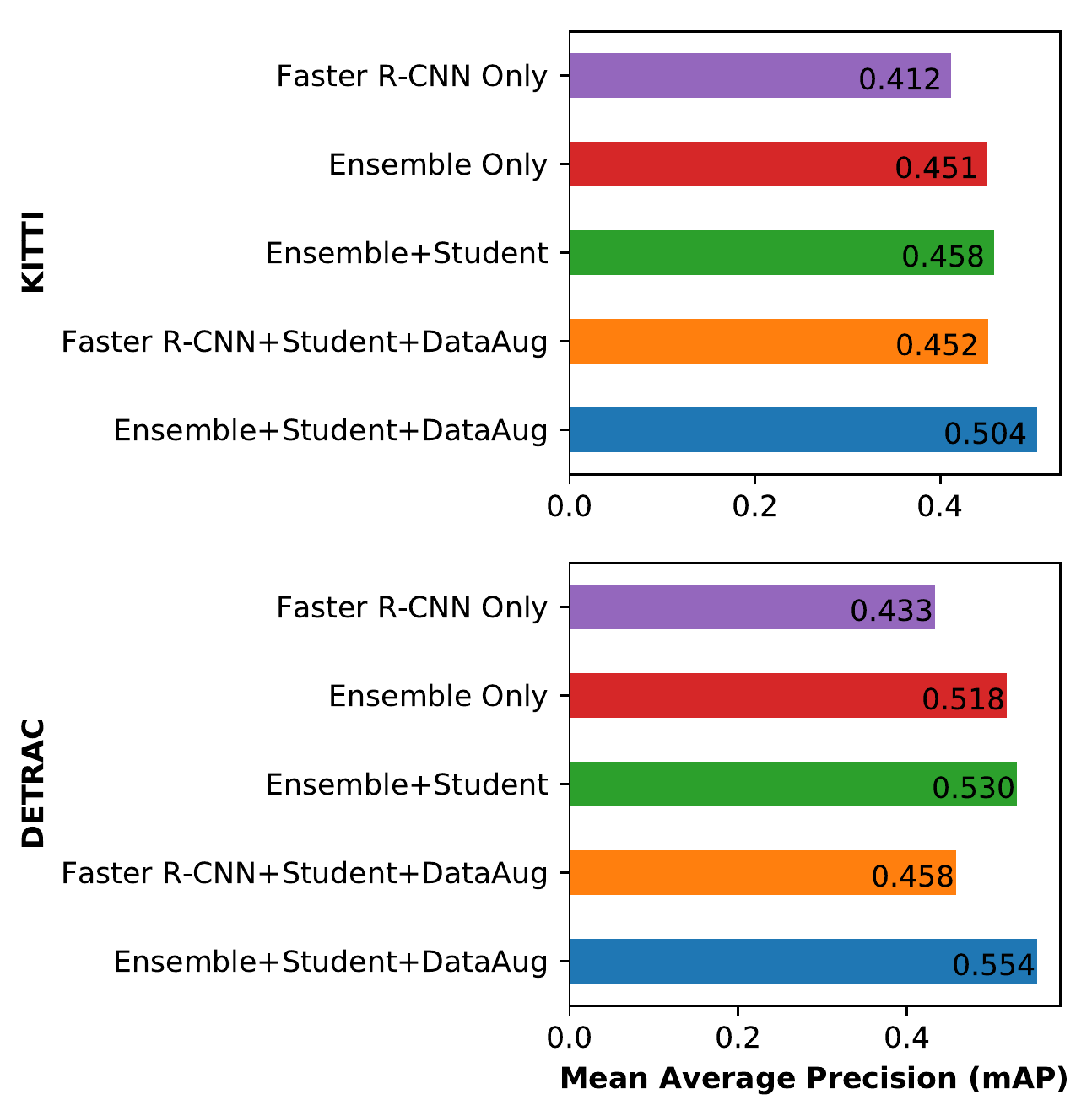}
    \caption{Ablation studies mAP scores.
             Without data augmentation, the Student Network achieves a comparable mAP
             score to the ensemble.
             We also show that it's possible to train a Student Network that's more
             accurate than its teacher using a single teacher.
    }
    \label{fig:ablation_studies}
\end{figure}

\paragraph{Pretrained Detector and Ensemble Baseline:}
Table~\ref{table:off_the_shelf_baseline} presents baseline detection
accuracies obtained using the pre-trained detectors.
The best pre-trained detector obtained a mAP score of 0.422 on KITTI and
0.458 on DETRAC.
Ensembling with coordinate and confidence averaging increases the mAP scores
to 0.436 and 0.471 respectively.
Finally, reweighting according to ensemble consensus further increases the
mAP\@ to 0.451 on KITTI and 0.518 on DETRAC (Figure~\ref{fig:accuracy_bars}).

\paragraph{Supervised Labeler Baseline:} We also present accuracies of the
Supervised Labeler, which is trained using labeled frames uniformly spaced
throughout each video and evaluated on the remaining frames (Figure~\ref{fig:label_plots}).
The Supervised Labeler is trained using the same architecture and training
parameters as the Student Network; it is also a MobileNet-SSD FPN detector
trained using data augmentation.
We evaluate the Supervised Labeler's predictions on all frames, and ensure that 
we override its predictions with ground truth on the labeled frames.

\paragraph{Mask R-CNN Comparison:} We also run Mask R-CNN~\cite{he2017mask},
an expensive but more accurate model.  Its accuracy is worse than either
the supervised baseline or the performance of ensembling+distillation,
but it does roughly match that of the reweighted ensemble alone
(Table~\ref{table:mask_rcnn}).
Were computational resources abundant, one could use 
Mask R-CNN as part of a stronger ensemble baseline.
Its high computational requirements suggest, however, that Mask R-CNN might
be more effective as an
``oracle'' that the system could query for frames where the ensemble is less
confident.
We observe that Faster R-CNN, YOLOv2, and SSD ResNet-50 FPN run at 5.3, 16,
and 6.4 FPS on a P100 GPU. Mask R-CNN runs at 1 FPS, which makes it 2.5 times
the cost of the ensemble, and 1.7x the cost of running the ensemble followed
by the distilled student.  On the small DETRAC and KITTI datasets, the training
cost of the student becomes dominant, but we hypothesize that for larger
videos, the training can be done with a subset of ensemble-labeled frames (but
necessarily defer exploring this 
future work for lack of a suitable fully labeled large video baseline).

\subsection{Improved Accuracy Through Multi-Source Knowledge Distillation}
\label{subsec:distill}
\begin{figure}
    \includegraphics[width=0.9\columnwidth]{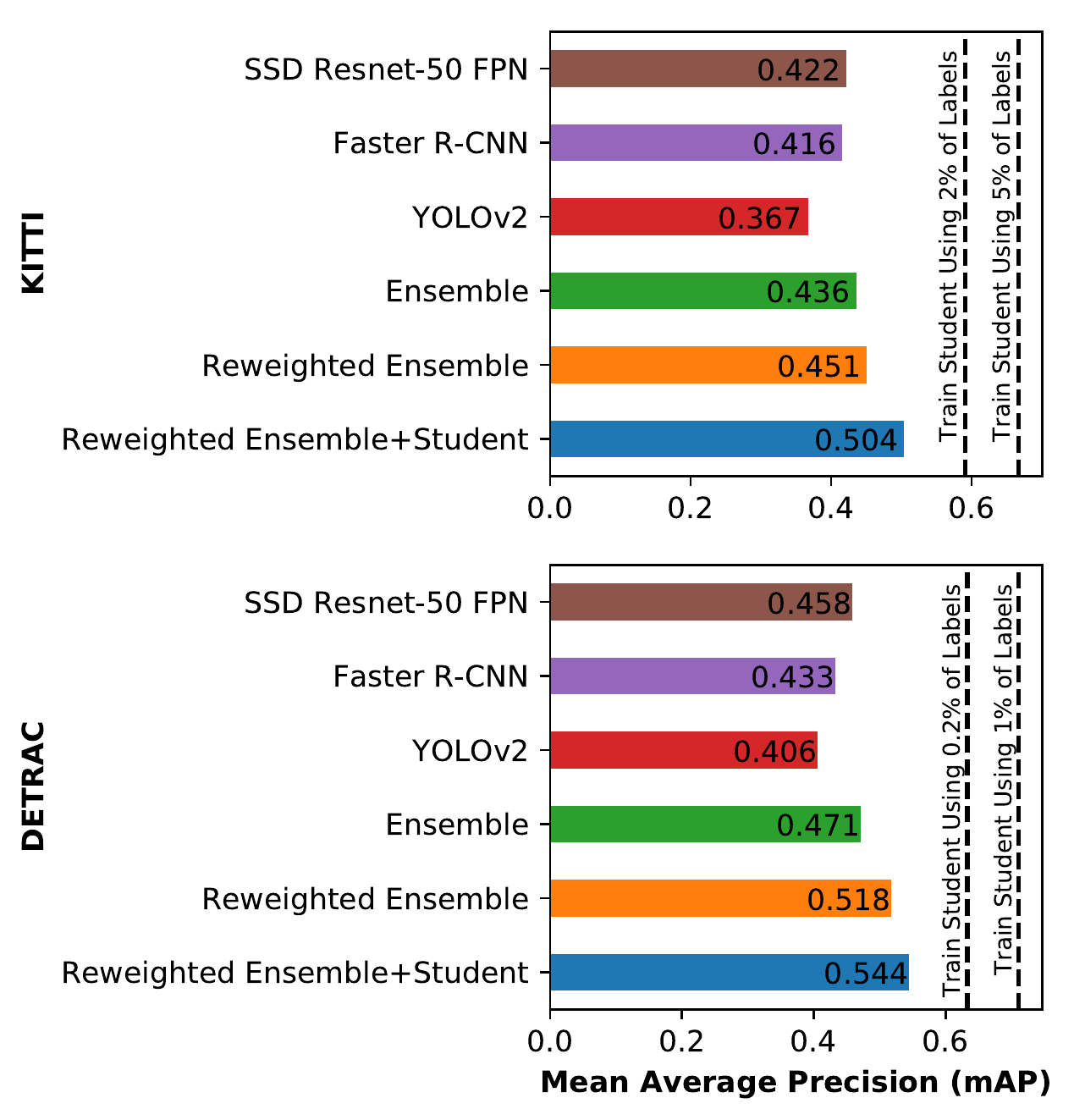}
    \caption{2\% and 5\% of KITTI corresponds to labeling 157 and 393 frames.
             0.2\% and 1\% of DETRAC corresponds to labeling 168 and 838 frames.}
    \label{fig:accuracy_bars}
\end{figure}

Training a MobileNet-SSD FPN Student Network on ensemble detections with data
augmentation improves mAP (compared to the ensemble) from 0.451 to 0.504 on
KITTI and from 0.518 to 0.554 on DETRAC (Figure~\ref{fig:accuracy_bars}).
There are several reasons why this may be the case.
First, while the ensemble is run
per-frame, the Student Network is trained on \emph{all} of the frames before it has to
make predictions about any of the frames, and therefore the training data
from other frames can affect its prediction on some frame $f$.
Second, data augmentation in the form of random cropping helps prevent the
Student Network from simply memorizing the data (the benefit of random cropping is examined in greater detail in Section~\ref{ablation}).
In addition, the similarity of nearby frames in video can also be thought of as
performing implicit data augmentation.

\begin{figure}

  \centering\includegraphics[width=0.95\columnwidth]{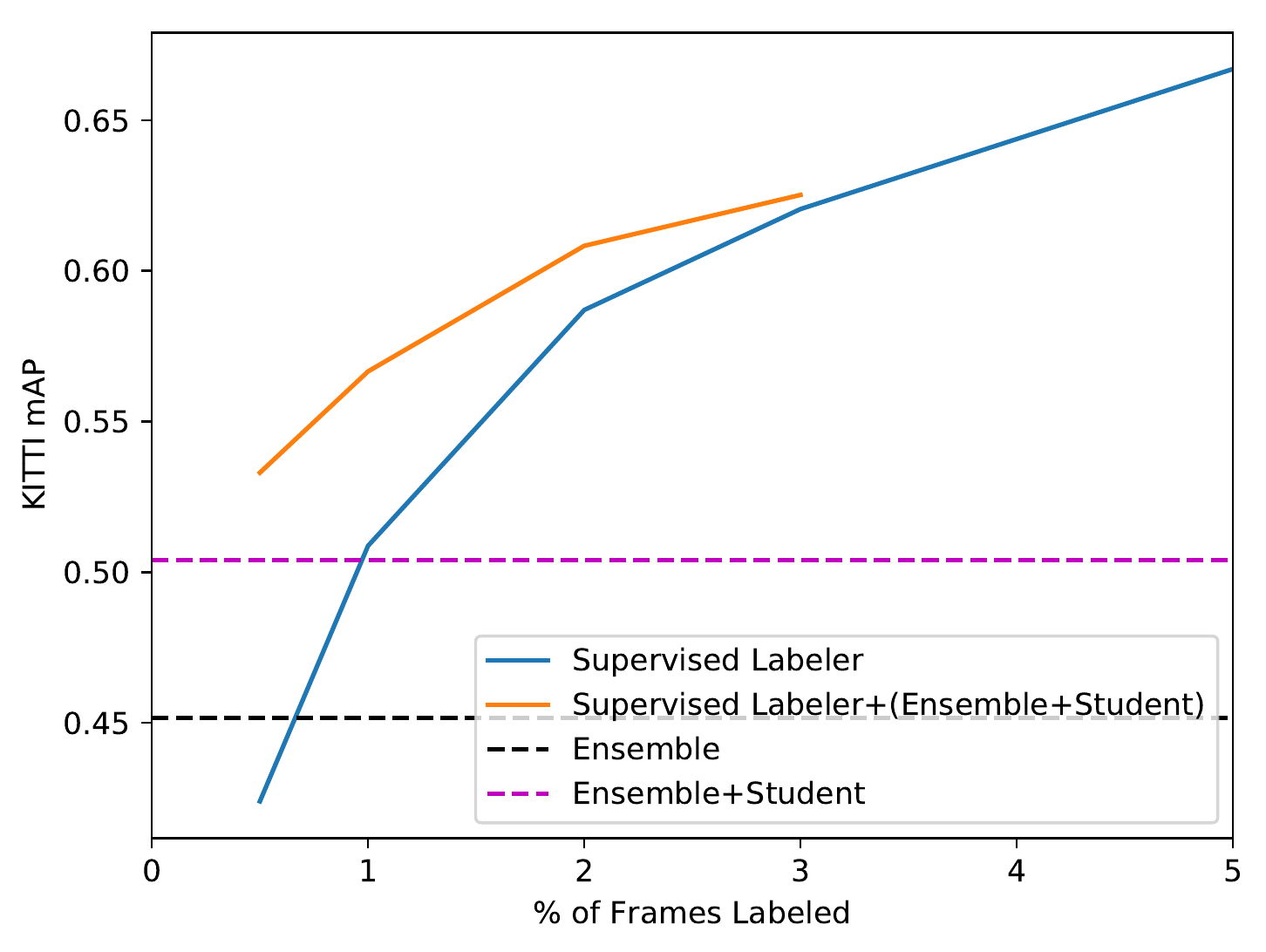}

    \centering\includegraphics[width=0.95\columnwidth]{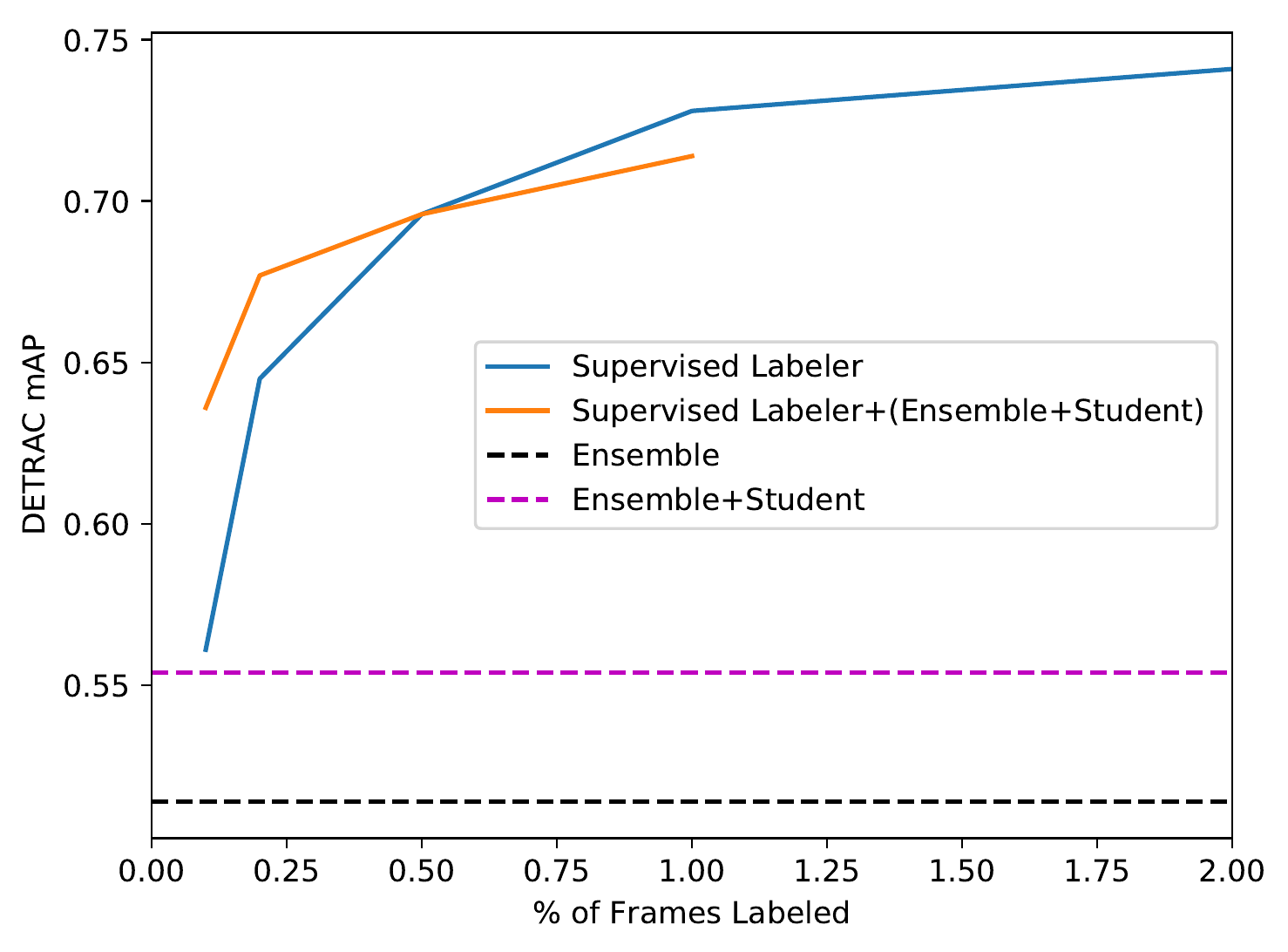}
    \caption{Combining Student Network detections with Supervised Labeler detections from
             leads to higher mAP until 3\% of KITTI
             (236 frames) and 0.5\% of DETRAC (419 frames) have been labeled.
             Note that the y-axis does not start from zero.}
    \label{fig:label_plots}
\end{figure}

Figure~\ref{fig:label_plots} shows that 
combining the Student Network detections with the Supervised Labeler detections greatly
improves mAP in the low-label regime compared to Supervised Labeler alone.
The Supervised Labelers trained on few labels perform poorly (compared to the
ensemble) because traditional transfer learning requires a sizeable number of
labels.
However, the Supervised Labelers still perform fairly well on what
may appear at first sight to be an astonishingly small number of labeled frames.
It is important to recall that these are fully annotated frames with all objects
labeled:
The crossover points (at 3\%
of KITTI and 0.5\% of DETRAC) of Figure~\ref{fig:label_plots} correspond to
labeling 992 bounding boxes across 236 frames on KITTI; and 2,771 bounding
boxes across 419 frames on DETRAC.
Lastly, we note that even when Supervised Labelers perform worse
than or comparably to
the Student Network, the fact that combining their detections significantly improves
mAP suggests that they are at least modestly complementary sources of knowledge.

\subsection{Ablation Studies}
\label{ablation}
In this section, we examine the individual contributions of several key pieces
of our work in greater detail. As shown in Figure~\ref{fig:ablation_studies},
using data augmentation, Feature Pyramid Networks, and an ensemble teacher
all help improve mAP.

\paragraph{The Effect of Data Augmentation:}
\label{data_aug}
Using data augmentation was crucial to training both
detectors to the desired level of accuracy.
Without data augmentation, the Supervised Labeler
frequently failed to converge, and its achieved accuracy was lower.
The Student Network also benefitted from
data augmentation:  Figure~\ref{fig:ablation_studies} shows that the Student Networks trained without
data augmentation only obtain mAP scores that are slightly higher than the ensemble mAP,
whereas applying data augmentation leads to a substantially
improved mAP\@.

\paragraph{Feature Pyramid Networks vs. Standard MobileNet-SSD:}
It was also important to use a sufficiently accurate base student architecture.
We initially used the standard MobileNet-SSD architecture, which (with data
augmentation) was able to nearly match the accuracy of the ensemble from which it
was trained.  We found, however, that the MobileNet-SSD Student Network
struggled with medium-sized occluded objects.
The Feature Pyramid Network-based detector, 
which is better at recognizing objects at different scales~\cite{lin2017feature}, handled these cases better,
while remaining sufficiently compact and fast.

\paragraph{Success Using a Single Teacher vs. An Ensemble:}
To understand whether our distillation technique is general enough to work
with a single teacher, we train a Student Network using only the
detections from Faster R-CNN. While the final mAP is lower than that of
training a Student Network with an ensemble teacher (as expected), distillation
from a single teacher improves mAP relative to that teacher to a similar
degree as it does when using an ensemble teacher.

\section{Discussion}
\label{discussion}

\paragraph{Having a strong baseline detector.}
Our work assumes a sufficiently strong baseline for the domain in question.
We use state-of-the-art, but fairly expensive object detectors on each
frame of the video collection.  Because we were able to use the Cloud TPUs
for training the student, but not for running inference, this step was,
surprisingly, the longest part of our pipeline.  We expect better results from running
ensembles of improved detectors, which may be reasonable when taking
advantage of optimized inference accelerators.

Our results show that the ensemble detectors trained on MS-COCO somewhat
generalize to scenarios where objects have relatively similar size and
aspect ratios.  They are unlikely, however, to work on drone and aerial
camera datasets.  For most common video domains, we believe there should
be a sufficient amount of labeled data that can be used to train a decent
general-purpose detector.

\begin{figure*}[thb]
    \centering
    \includegraphics[width=0.25\textwidth]{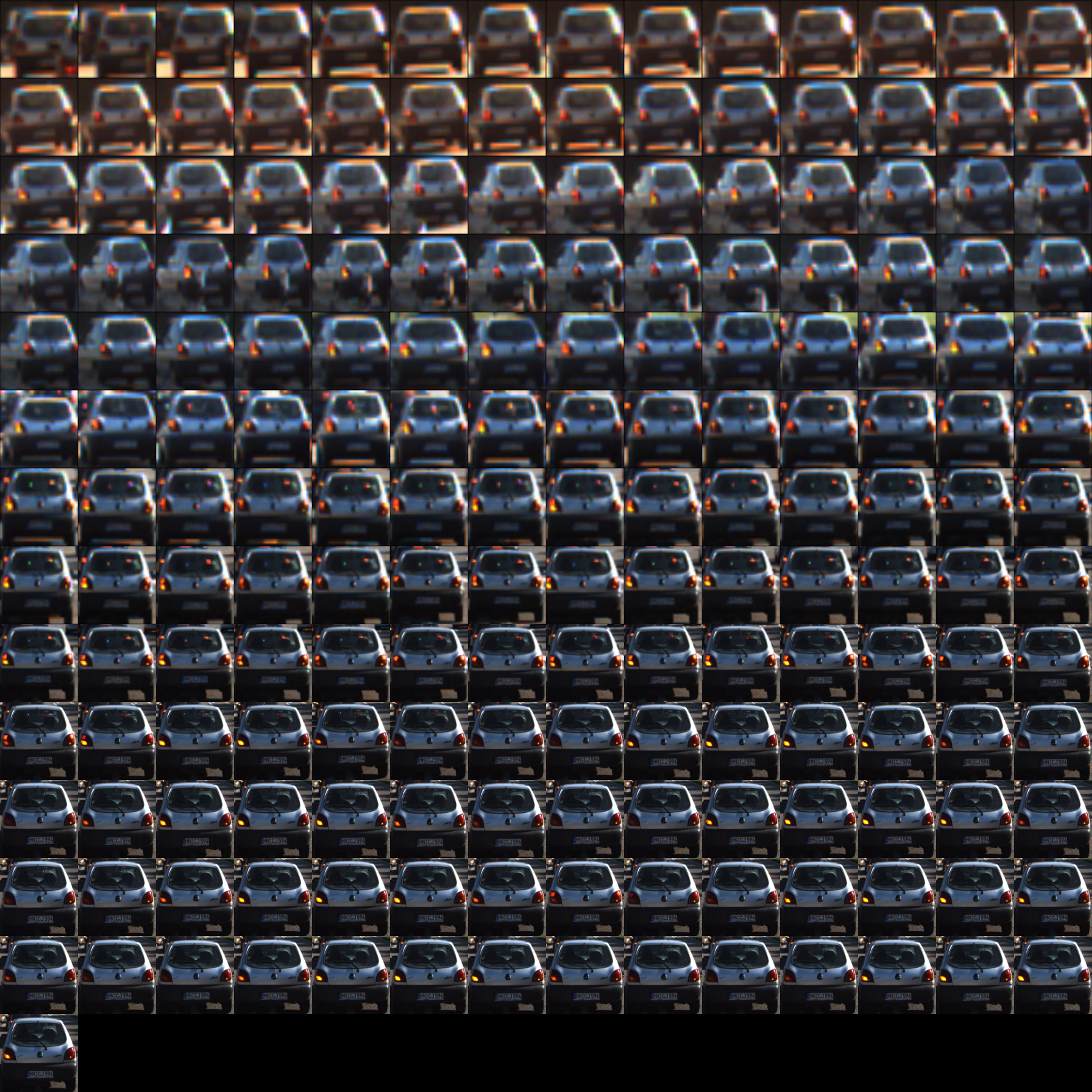}
    \includegraphics[width=0.33\textwidth]{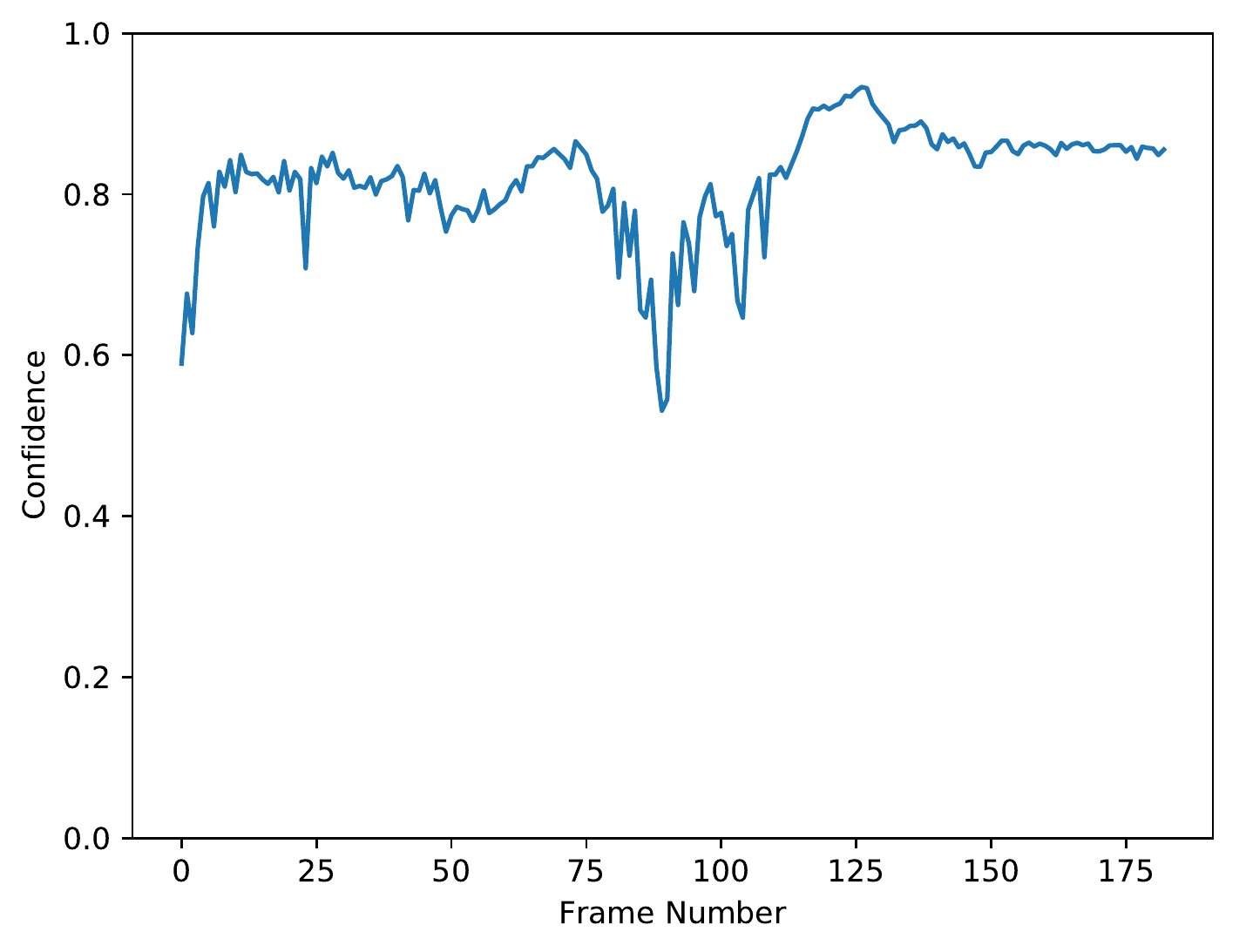}
    \includegraphics[width=0.33\textwidth]{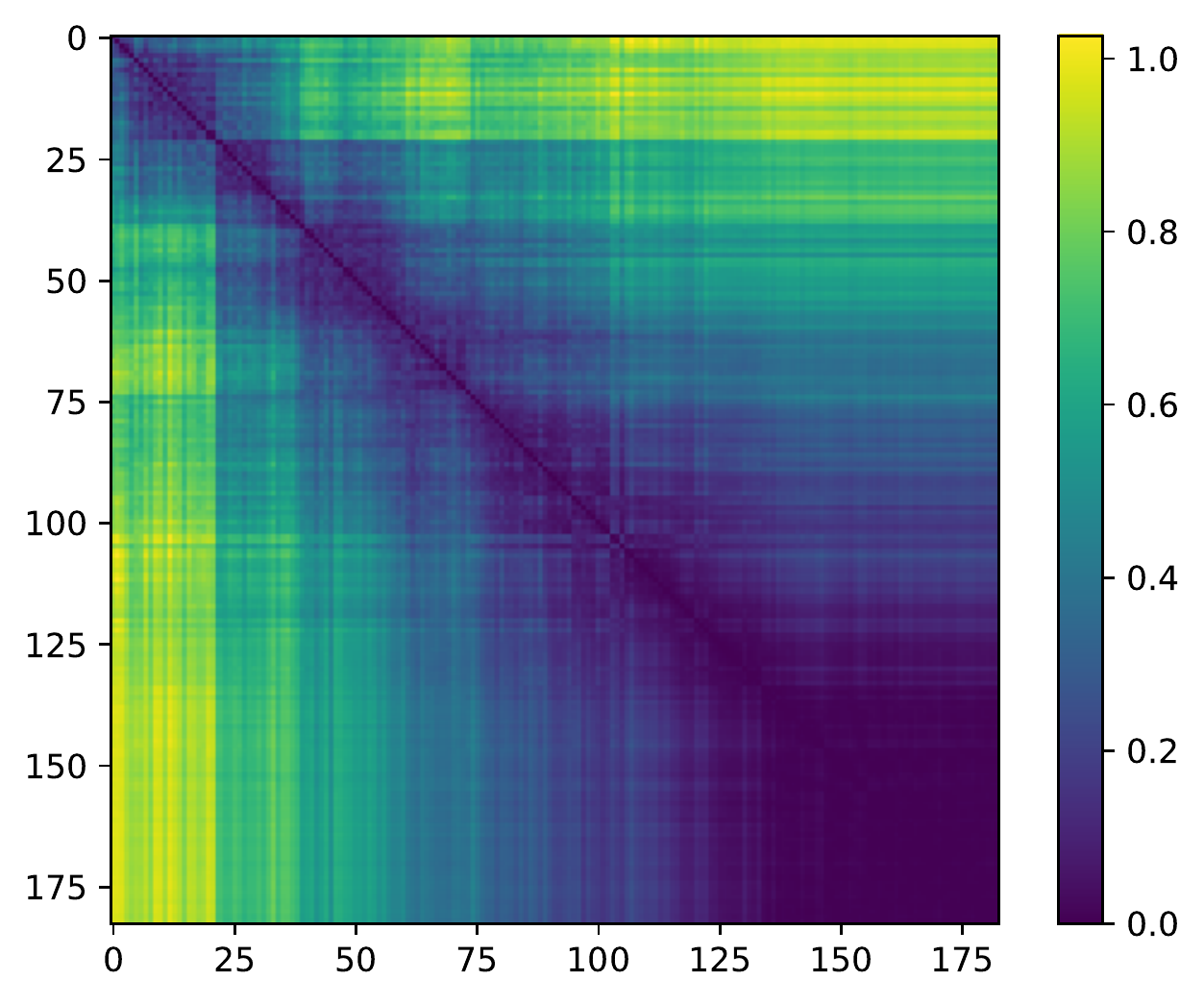}
    \caption{Tracked car sequence exiting a highway in KITTI. There is a sharp
             drop in confidence at the halfway through the sequence (likely due
             to the rapidly changing surroundings) despite the car's appearance
             remaining relatively consistent, as indicated by the lack of
             sharp discontinuities in the center of the pairwise cosine
             distance matrix.}
    \label{fig:unstable_confs}
\end{figure*}

\paragraph{When to label frames of a video:}
Given a fixed labeling budget $B$ significantly smaller than the total number
of frames $N$, an ideal approach would be to acquire labels for a maximally
diverse set of frames.
In this paper, we do not address the question of how to
choose the best $B$ frames to label. We simply label $B$ evenly spaced frames
throughout a video and discuss how to choose $B$ based on how long objects
persist in a video.

Given frequent-enough annotations, existing video annotation tools can track and
interpolate object
bounding boxes across frames. Using our uniform labeling approach, we can think
of there being a minimum labeling interval where it becomes possible to recover
the labels for all of the frames in the video. This ``phase-transition''
point is different for each video.
We examine the median object durations in KITTI and DETRAC as a proxy for the
rate of change, and find them to be 32 and 71 frames respectively.
We thus choose (1/20=5\%) and (1/50=2\%) as the highest labeled baseline for
each respective dataset.

\paragraph{Different detectors contribute different biases to the training:}
Each of the ensemble detectors is pre-trained on the same dataset yet
each one contributes to the student's accuracy because they have
different biases.
We were surprised by the degree to which the detectors varied.
For instance, when running on KITTI, Faster R-CNN produced the
most detections with high confidence, yet SSD ResNet-50 FPN was actually more
accurate than Faster R-CNN.
Our work treats all ensemble detectors equally and could be improved by
examining the correlations between the ensemble members in greater detail.
Right now, we require fairly strong ensemble detectors, but properly accounting
for inter-detector correlations could enable us to construct larger
ensembles of weaker detectors.

\section{Future Work}

Our core plans for future work are to incorporate additional video-derived
constraints into improving the labels fed to the student network.

Manual inspection of mistakes made by the off-the-shelf detectors on
a subset of frames in KITTI suggest that heavy occlusion is one of the
major sources of error for the off-the-shelf detectors, and that
our techniques help correct some of them.  This suggests that
approaches that can improve bounding box detection when an object
is (often transiently, in video) occluded may provide a rich source
of accuracy improvements.

One such source, which can also correct for transient detection failures,
is to incorporate object tracking to 
boost the accuracy of object detections by the ensemble.
Object tracking and metric learning associate detections across
frames, smoothing out detector confidences over time.
Previous work shows that slight modifications to a frame can have significant
non-local
impacts on object detection~\cite{rosenfeld2018elephant}.  In video, this
manifests as detections of the same object having unstable confidence and
localization across frames (Figure~\ref{fig:unstable_confs} shows an example
from the KITTI dataset).  We believe
that object tracking holds promise for improving detection confidence.
Multiple Object Tracking (MOT)~\cite{milan2016mot16} and related
challenges provide the basis for 
identifing objects with
\emph{appearance constancy}---those that maintain a similar
appearance over time---to normalize the ensemble's prediction
confidence across the stream.
Our preliminary approaches to do using off-the-shelf
tracking methods (Kalman filters, correlation filters, and regression-based
trackers) were thus far unsuccessful, but we plan to return to this
approach in the future.

\section{Conclusion}

This paper presented a new method for using an ensemble of object
detectors, together with in-domain distillation to a Student Network,
that outperforms all members of the ensemble and the aggregate ensemble
itself.  The presented method can further be combined with small numbers
of labeled training samples to increase accuracy above either the ensemble
or Supervised Labeler.  Despite using  weaker detectors in its
ensemble, the ensemble and student together outperform state of the art
methods such as Mask R-CNN, resulting in a practical method for 
for unlabeled (or lightly-labeled) object detection on video.

{\small
\bibliographystyle{ieee}
\bibliography{paper}
}

\end{document}